\documentclass{article}
\usepackage{spconf,amsmath,epsfig}
\usepackage{hyperref}
\usepackage[backend=bibtex, sorting=none]{biblatex}
\usepackage{multicol}
\usepackage{xcolor}
\usepackage{adjustbox}
\usepackage{amssymb}
\usepackage{booktabs}
\usepackage{color}
\usepackage{multirow}
\usepackage{graphicx}
\usepackage{wrapfig}
\usepackage{bm}
\usepackage{pifont}

\title{FANet: Feature Amplification Network for Semantic Segmentation in Cluttered Background}

\name{Muhammad Ali$^1$, Mamoona Javaid$^2$, Mubashir Noman$^1$, Mustansar Fiaz$^3$, Salman Khan$^1$
}
\address{$^1$MBZUAI, UAE \hspace{1.5mm} $^2$Institute of Space Technology, Pakistan \hspace{1.5mm} $^3$IBM Research
}

\begin{document}
\maketitle
\begin{abstract}

Existing deep learning approaches leave out the semantic cues that are crucial in semantic segmentation present in complex scenarios including cluttered backgrounds and translucent objects, etc. To handle these challenges, we propose a feature amplification network (FANet) as a backbone network that incorporates semantic information using a novel feature enhancement module at multi-stages. To achieve this, we propose an adaptive feature enhancement (AFE) block that benefits from both a spatial context module (SCM) and a feature refinement module (FRM) in a parallel fashion. SCM aims to exploit larger kernel leverages for the increased receptive field to handle scale variations in the scene. Whereas our novel FRM is responsible for generating semantic cues that can capture both low-frequency and high-frequency regions for better segmentation tasks. We perform experiments over challenging real-world ZeroWaste-f \cite{bashkirova2022zerowaste} dataset which contains background-cluttered and translucent objects. Our experimental results demonstrate the state-of-the-art performance compared to existing methods. The source code can be found at \url{https://github.com/techmn/fanet}.

\end{abstract}


\section{Introduction}
\label{sec:intro}
Semantic segmentation is a fundamental computer vision task having objective to obtain pixel-level predictions and is critical for various practical applications such as autonomous driving, scene understanding, robot sensing, and image editing. 
Traditional convolutional neural networks (CNNs)-based methods rely on local short-range structure to capture the semantics of the image \cite{long2015fully, fiaz2019convolutional}. However, due to intrinsic fixed geometric structures, they are limited to short-range contextual information. Various efforts have been made to tackle this issue including dilated convolution \cite{wang2018understanding, wu2019fastfcn, gao2023rethinking} and channel or spatial attention models \cite{huang2022channelized, chen2021channel}. 

\begin{figure}[t]
   \centering
    \includegraphics[width=\linewidth]{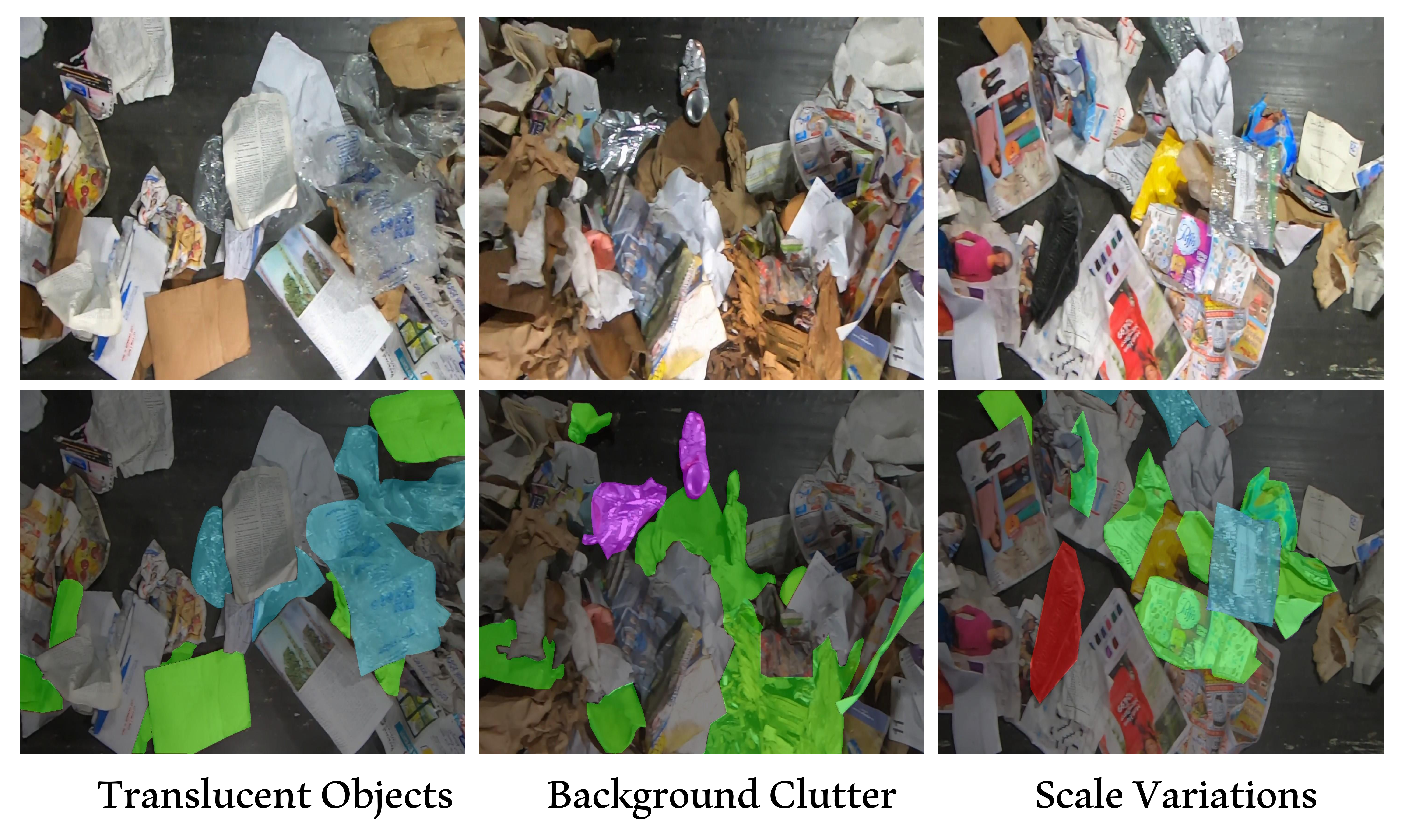}
    \caption{The main challenges in semantic segmentation, e.g., translucent objects, background clutter, and scale variations. The first row indicates the input image while the bottom row shows the input image overlay with the ground truth.}    
   \label{fig_intro_challenges}
\end{figure}

Since the advent of vision transformers \cite{dosovitskiy2020image} which make use of the self-attention mechanism, semantic segmentation has been considerably revolutionized.
For instance, \cite{xie2021segformer,shi2023transformer} effectively utilizes the global contextual relationships of the transformers to perform reasonably on semantic segmentation tasks.
However, they pay less attention to capturing the semantic-level contextual information. In addition, the dominance of global receptive fields and long-range dependencies hinder its ability to capture the fine details. 
Moreover, attention-based models are data-hungry and require a large amount of data for optimal convergence due to the lack of inductive bias and the struggle to capture local dependencies due to dominant global representations which restrict their practicality. To this end, several hybrid attention models \cite{wu2021cvt, wang2023internimage, hatamizadeh2023global, noman2024elgc, fiaz2023sat} are introduced to improve the performance by utilizing convolutions and self-attention mechanisms. However, these models still face difficulties to perform better when the segmentation objects lie in a cluttered background. Furthermore, the segmentation of translucent objects is another challenging task that needs special consideration. 
As shown in Fig. \ref{fig_intro_challenges}, these models may suffer from three challenges. (i)- The model may encounter notable difficulties when dealing with translucent objects due to their intrinsic nature of opacity and unclear boundaries between object and background. (ii)- Background clutter makes appearance representations more ambiguous. (iii)- Diverse scale variations increase the difficulty of capturing subtle objects.

\begin{figure*}[t!]
   \centering
    \includegraphics[width=\linewidth]{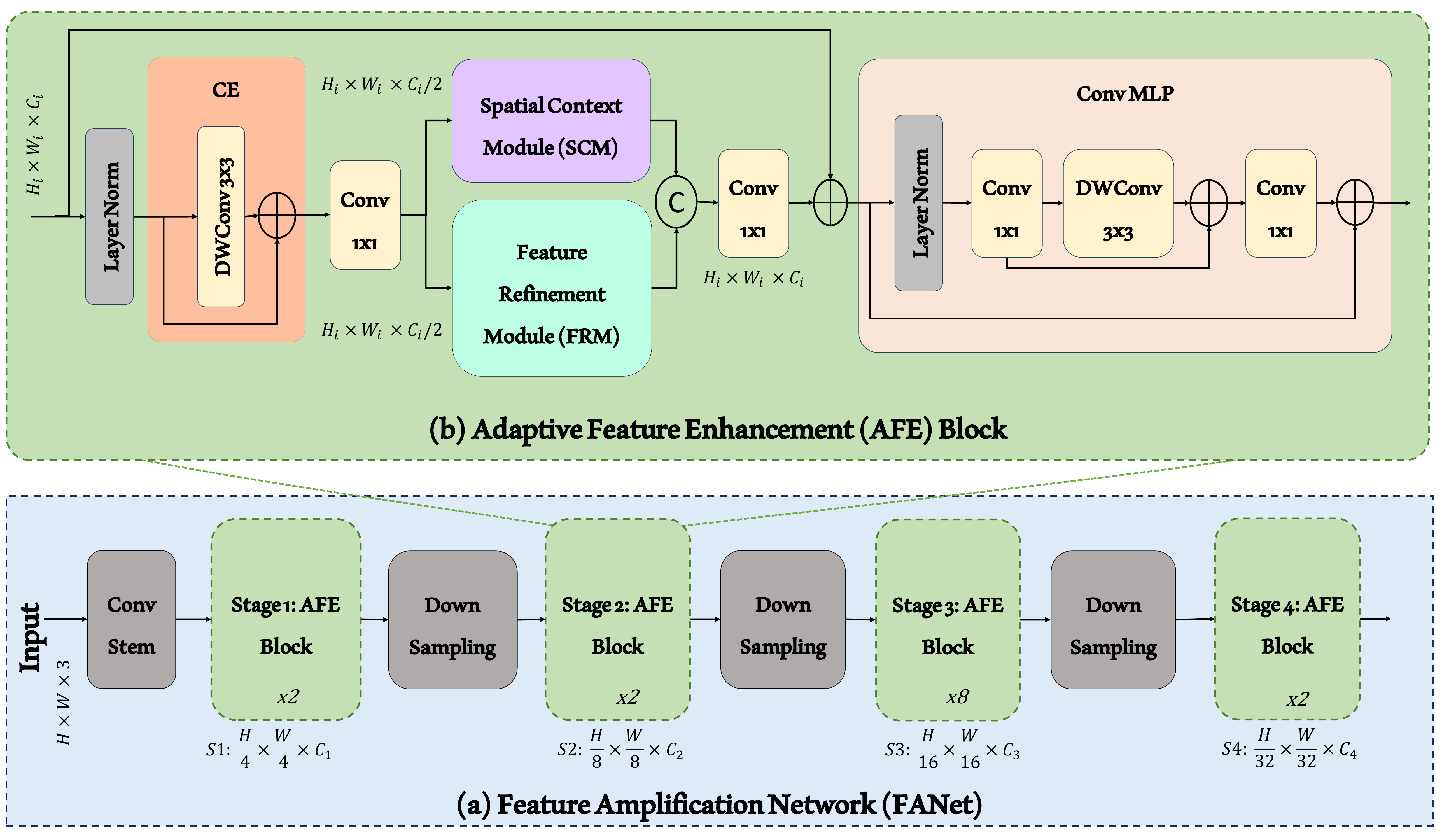}
    \caption{The (a) is the overall illustration of our proposed feature amplification network (FANet), as a backbone network, for background-cluttered semantic segmentation.  The input is passed to a backbone network  (FANet) to produce the multi-stage features ($S1, S2, S3,$ and $S4$). These multi-stage features are input to a UperNet decoder \cite{xiao2018unified} as a segmentation head for prediction.  The (b) shows our novel adaptive feature enhancement (AFE) block. Our (b) AFE block is designed to capture the rich information. It comprises convolutional embeddings (CE), spatial context module (SCM),  feature refinement module (FRM), and ConvMLP.  Our AFF block adaptively aggregates the large kernel information using SCM which increases the receptive field and FRM which refines the features in the spatial dimension. }    
   \label{fig_proposed_framework}
\end{figure*}

\noindent\textbf{Contributions:} In this work, to handle the aforementioned challenges for the semantic segmentation task, we propose a feature amplification network (FANet) as shown in Fig. \ref{fig_proposed_framework}-(a).
We introduce FANet as a backbone to capture the enhanced features and generate multi-stage features using our novel adaptive feature enhancement (AFE) block, as shown in Fig. \ref{fig_proposed_framework}-(b). 
Our plug-and-play AFE block aims to benefit from a larger kernel and semantic cues in a \textit{parallel} manner, which can extract more comprehensive features to preserve the coarse-to-fine details so that the boundaries of the objects are easily distinguishable.
The focus of the design is to encode the intrinsic properties of the target objects and segment the objects from the complex scenes, especially in the presence of cluttered backgrounds.
Our block is responsible for explicitly exposing the spatial descriptors while simultaneously performing feature enhancement to preserve both high-frequency and low-frequency components in an image. To excavate the spatial context, we propose a spatial context module (SCM) that uses a large kernel to handle the scales of the objects in complex scenes.
Meanwhile, to inscribe the semantic cues, we introduce a novel feature refinement module (FRM) which aims to capture low-frequency components as well as highlight the fine details of the objects.
Experimental results indicate a considerable improvement in the complex ZeroWaste-f \cite{bashkirova2022zerowaste} dataset compared to the existing state-of-the-art methods.

\begin{figure}[t]
   \centering
    \includegraphics[width=1.0\linewidth]{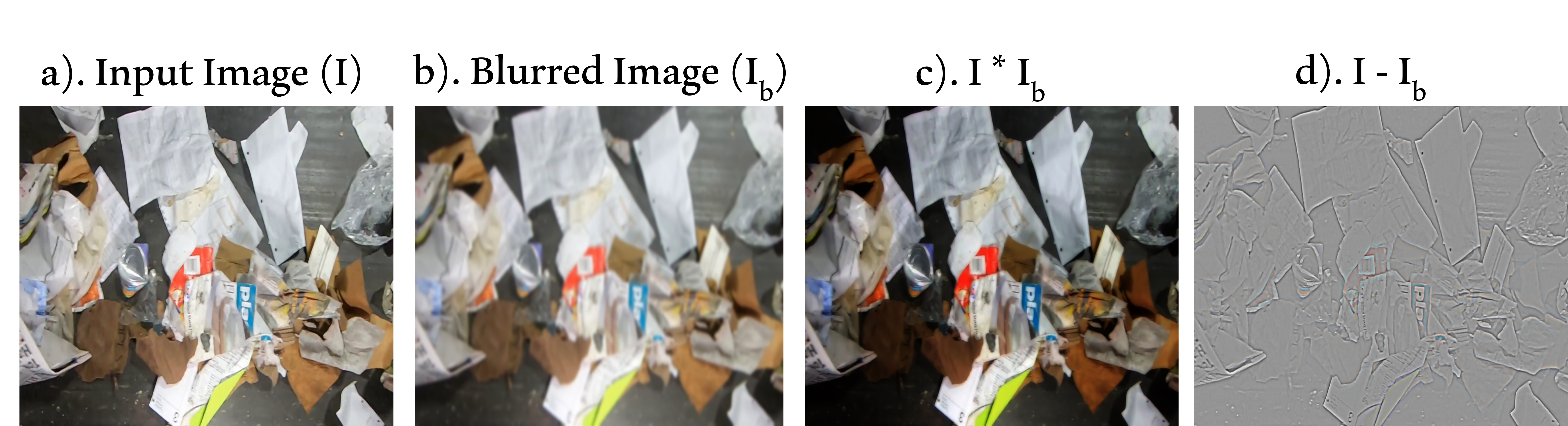}
    \caption{ Given an image $I$ (a) and its smoothed version $I_b$ (b), element-wise multiplication of $I$ and $I_b$ emphasize the color information and preserve the blob regions (c). Whereas (d) fine details can be highlighted by subtracting the smoothed image $I_b$ from the original image $I$. Motivated by this, we introduce our feature refinement module (FRM).  }
    \vspace{-1.0em}
    \label{fig:teaser_figure_showing_enhancement_idea}
\end{figure}

\section{Background}
Typically, 
image sharpening and contrast enhancement are the two fundamental concepts that are applied in the spatial domain to improve the quality of the image, as shown in Fig. \ref{fig:teaser_figure_showing_enhancement_idea}. The sharpening is performed to emphasize the high-frequency or fine details in a given visual feature. To do so, the visual feature is subtracted from the Laplacian function to obtain a sharpened result. Suppose, $f(x,y)$ is the given input feature then the resultant sharpened feature $g(x,y)$ is obtained using Laplacian follows:
\begin{equation}
\label{eq:sharp}
    g(x,y) = f(x,y) - c[\Delta^2 f(x,y)],
\end{equation}
where $c$ is the Laplacian mask center coefficient which is empirically determined.  

Contrast enhancement is another image enhancement technique that improves the contrast in a feature by stretching the range of intensity values to span a desired range of values to highlight the low-frequency regions in the feature map. 
The contrast-enhanced feature can be obtained as:

\begin{equation}
\label{eq:contrast}
    q(x,y)= f(x,y)\odot m(x,y),
\end{equation}

where $q(x,y)$ and $\odot$ represent the output feature and Hadamard product operation, respectively. The $m(.)$ is designed to stretch the contrast and enhance features and can be defined as follows:

\begin{equation}
    m(x,y) = \gamma (\frac{1}{1+e^{-\alpha((x,y) - \beta}}) - 0.5),
\end{equation}

where $\gamma$ is a scaling factor used to control the strength of the enhancement. The $\alpha$  and $\beta$ control the contrast of the enhancement function.

Finally, the outputs from Eq. \ref{eq:sharp} and Eq. \ref{eq:contrast} can be combined to get the enriched features in the spatial domain that can capture both high-frequency and low-frequency regions.  
Motivated by this, we leverage our feature amplification module described in Sec. \ref{sec:fam_details} for token-mixer design. 

\section{Method}

\subsection{Overall Architecture}  \label{sec:overall}
Fig.~\ref{fig_proposed_framework}-(a) shows the overall architecture of our proposed feature amplification network (FANet) for the semantic segmentation task. 
The focus of our design is a backbone network that can capture the intrinsic properties of the object and segment it from the cluttered background, using our novel adaptive feature enhancement (AFE) block. Specifically, our novel FANet can capture the enriched features and generate multi-stage features ($S1, S2, S3,$ and $S4$). To obtain the multi-stage features, the input $ x \in \mathbb{R}^{H \times W \times 3}$ is input to non-overlapping convolution stem layers (kernel size = 5×5, stride = 4) to generate the tokens of size $ \mathbb{R}^{H/4 \times W/4}$. Following hierarchical design \cite{yang2022focalnet,rao2022hornet, liu2021swin}, our FANet comprises four stages to obtain hierarchical feature representations. There exists a down-sampling convolution layer (kernel size = 3×3, stride = 2) between two stages to reduce the spatial resolution of the features. The multi-stage features are input to a UperNet \cite{xiao2018unified} decoder to get the final segmentation mask.  

\subsection{Adaptive Feature Enhancement (AFE) Block} \label{sec:afe_block_details}
The proposed AFE block comprises four key components: convolutional embeddings (CE), spatial context module (SCM), feature refinement module (FRM), and convolutional multi-layer perceptron (ConvMLP). The focus of this design is to adaptively capture the enriched features of cluttered backgrounds for semantic segmentation. The input features are passed through a LayerNorm and a CE to learn the generalization and discriminative ability \cite{ wu2021cvt}. The output of CE is passed to a 1x1 convolution layer that squeezes the channels to half. The channel squeeze helps to reduce the computational overhead and encourages the model to perform feature mixing. The squeezed features are passed to the SCM module which comprises group-wise convolution with a larger kernel (kernel size = 7x7). The objective of SCM is to increase the receptive field which can capture the spatial context over a broader range to handle scale variations. In parallel, the squeezed features from the CE layer are also fed to the feature refinement module (FRM) to refine the features (see details in Sec. \ref{sec:fam_details}). The outputs from SCM and FRM are fused and projected through a 1x1 convolution layer and a ConvMLP to further enhance the representations.

\subsection{Feature Refinement Module (FRM)} 
\label{sec:fam_details}
\begin{figure}[t]
   \centering
    \includegraphics[width=\linewidth]{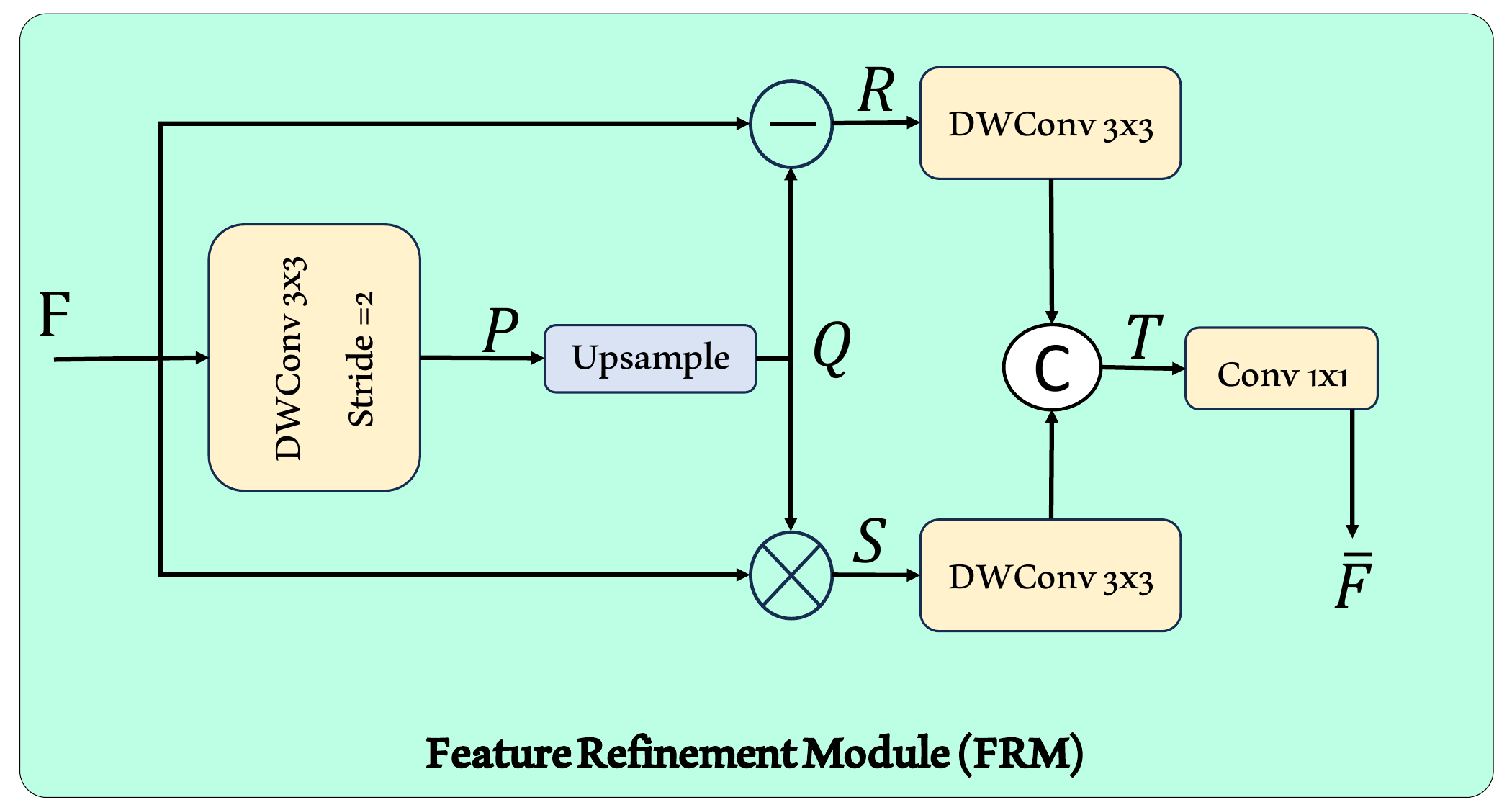}
    \caption{The illustration of our novel feature amplification module. The input features $F$ are downsampled using depthwise convolution (DWConv) and upsampled to get $Q$ features. The input features $F$ are subtracted from $Q$ features to get $R$ features that highlight the fine details. Similarly, the input features are multiplied with $Q$ features to obtain $S$ features which highlight the low-frequency components in the spatial dimension. Later, these low-frequency and high-frequency features are aggregated after DWConv to obtain enhanced features. Finally, the aggregated features are input to the projection layer to obtain the final $\bar F$ features.}    
   \label{fig_fam_module}
\end{figure}

Inspired by the image sharpening and contrast enhancement concept, we introduce a novel feature refinement module (FRM) to capture the low-frequency context and emphasize the blob regions while simultaneously highlighting the high-frequency details. The block diagram for FRM is depicted in Fig.~\ref{fig_fam_module}. Suppose  $F \in \mathbb{R}^{C \times H \times W}$  be the input of FRM. We pass it from a depthwise convolution layer to obtain down-sampled feature maps $ P \in \mathbb{R}^{C \times H/2 \times W/2}$. In the image processing domain, high frequencies are highlighted by taking the difference between the image and its blurred version. We adopt a similar procedure by up-sampling the smoothed feature maps $P$ to the same spatial resolution of $F$ to obtain $Q$ features. Later, the difference between  $F$ and $Q$ highlights the fine details resulting in $R$ refined feature embeddings (as shown in Fig.~\ref{fig_fam_module}). 

The second branch (bottom) in FRM strives to capture the low-frequency regions in the feature maps. Given a normalized image, the blob features can be emphasized by element-wise multiplication of the image with its blurred version. To do so, we perform element-wise multiplication operations between the $F$ and $Q$ features to obtain $S$ which capture the low-frequency components. The idea is to highlight the low frequencies by focusing on the blobs in the feature maps. Our FRM emphasizes the low and high-frequency regions and concatenates them after depthwise convolution in the channel dimension (to obtain $T$).  Finally, these features are realized with a projection layer to obtain the final enriched/amplified features $\bar F$.

\section{Experimentation}

\begin{table}[t!]
\centering
\caption{Comparison of ours FANet with state-of-the-art methods for semantic segmentation on the test set of the  ZeroWaste-f dataset. We report the results in terms of mIoU and pixel accuracy (Pix. Acc.). Here, the FocalNet-B provides improved performance compared to the DeepLabv3+. Our FANet when utilized as a backbone network improves mIoU performance by 1.63\% over the FocalNet-B. The best results are in bold.}
\scalebox{1.2}{
\begin{tabular}{|l|c|c|}
\hline
\textbf{Methods} & \textbf{mIoU} & \textbf{Pix. Acc.} \\
\hline
CCT \cite{9157032}   &  29.32 & 85.91 \\
ReCo \cite{ReCo}   & 52.28  & 89.33 \\
DeepLabv3+ \cite{chen2017deeplab}  &  52.13 &  {91.38} \\
FocalNet-B \cite{yang2022focalnet}   & 53.26 &{91.28} \\
\hline
FANet (ours) & \textbf{54.89} & \textbf{91.41} \\
\hline
\end{tabular}
}
\label{tab:comparison_of_fanet_on_zerowaste}
\end{table}

\begin{figure*}[t!]
   \centering
   \includegraphics[width=\linewidth]{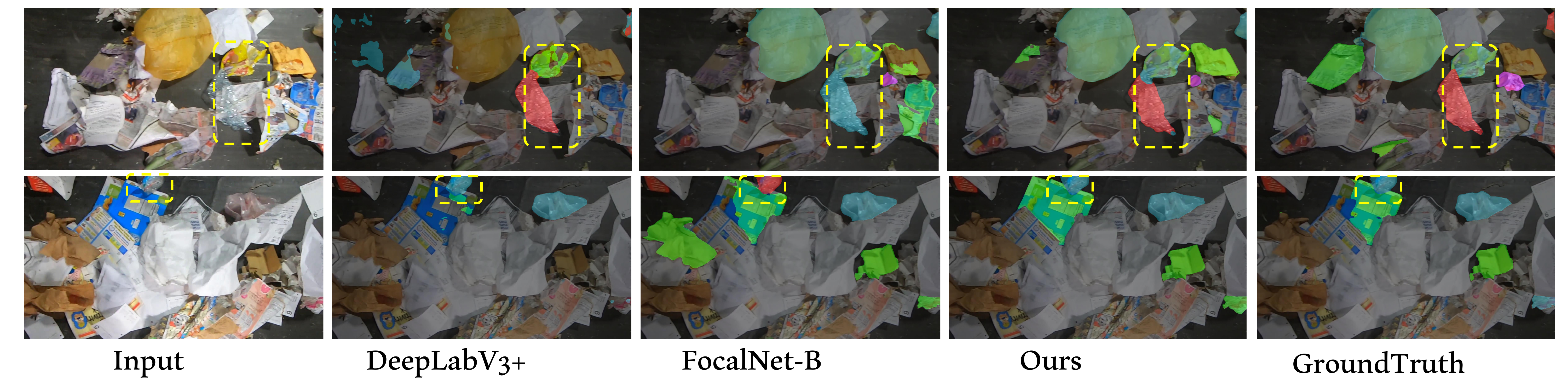} 
    \caption{Qualitative comparison on   ZeroWaste-f dataset. Our method better segments the objects from cluttered backgrounds compared to existing state-of-the-art methods. }
    \label{fig_qualitative}
\end{figure*}

\subsection{Dataset and Evaluation Protocol}
We have chosen to focus on the   ZeroWaste-f dataset \cite{bashkirova2022zerowaste}
which is introduced to advance the field of waste management and segmentation, particularly in environments with extreme clutter, while also boosting innovations in recycling applications. It encompasses a diverse range of waste types, including various paper, cardboard, plastics, metals, and organic materials, often in cluttered and overlapping states. 
Our motivation in selecting this dataset is twofold: firstly, to push the boundaries of what's possible in the realm of automated waste segmentation, and develop a model that can accurately differentiate and classify a wide array of waste materials. Secondly, we aim to contribute to environmental sustainability efforts. By improving the precision of waste segmentation, we can enhance recycling processes and waste management practices, ultimately contributing to a more sustainable and environmentally conscious approach to waste disposal. 
This dataset comprised 3002 training, 572 validation, and 929 test sets.  
Following \cite{bashkirova2022zerowaste}, we 
use the mean intersection over union (mIoU)  and pixel accuracy to measure the performance of the models. The mIoU is a crucial metric for assessing segmentation accuracy by quantifying the overlap between prediction and ground truth.

\subsection{Implementation Details}
We implemented our method in PyTorch and trained over NVIDIA V100 GPU.
For this purpose, we leverage the MMSegmentation open-source toolbox \cite{contributorsmmsegmentation}. The training of our model is conducted using the UperNet architecture, with the backbone initialized using weights pre-trained on the ImageNet-1K \cite{deng2009imagenet} dataset. This model undergoes a training regimen of 40k iterations on the ZeroWaste dataset. We employ the AdamW optimizer \cite{kingma2017adam} for this process, starting with an initial learning rate of $9e^{-5}$.
In line with standard research methodologies, our training approach includes a learning rate decay, specifically utilizing a polynomial decay strategy with a power exponent of 1.0. We configure our training setup with a crop size of 512x512 and a batch size of 2. 
Following \cite{liu2021swin, guo2023visual},
we first trained the backbone on the ImageNet-1K  \cite{deng2009imagenet} dataset for 300 epochs with batch size 80 using 16 NVIDIA V100 GPUs and a learning rate of 1e-3. We adopt the AdamW optimizer 
\cite{loshchilov2017decoupled}  for training and set weight decay of
$0.05$.
After pertaining, we finetuned the model on the semantic segmentation dataset.

\begin{figure}[t]
   \centering
    \includegraphics[width=1\linewidth]{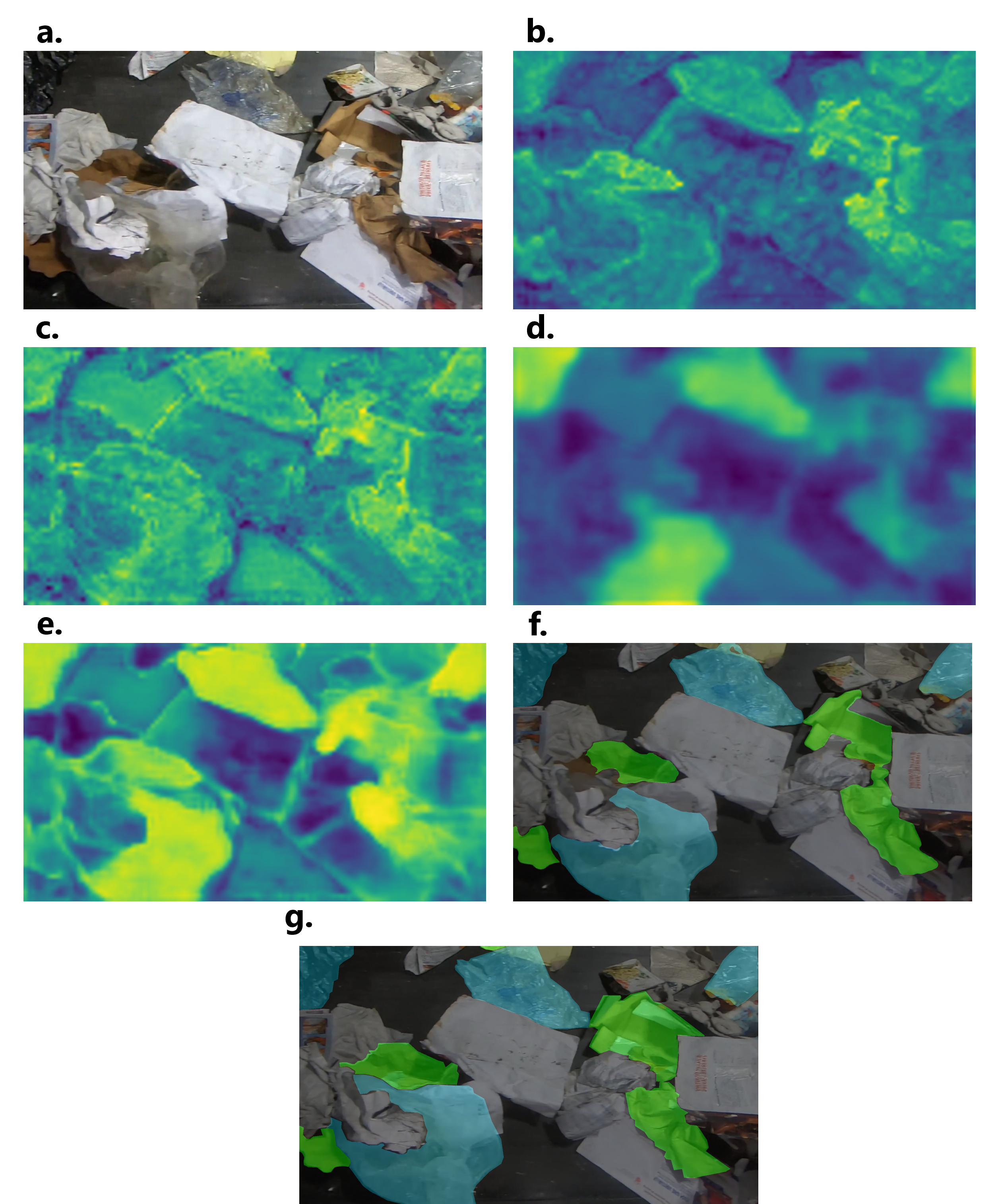}
    \caption{Illustration of the impact of feature refinement module (FRM). (a) is the input to the model, (b) shows the output feature maps of the third stage and input to the FRM, (c) and (d) show the highlighted features in the FRM as $R$ and $S$, respectively. The (e) shows the enhanced features as output provided by FRM, and (f) highlights the resulting segmentation map provided by the model overlay with the input image. The (g) is the ground truth overlay with the input image.}    
   \label{teaser_figure_showing_FAM}
\end{figure}

\begin{table*}[t!]
\centering
\caption{Ablation study of our FANet pretrained over ImageNet-1K for 300 epochs. First, we use the depthwise convolutional embeddings using a skip connection called CE and ConvMLP as our baseline. Row 2 shows that introducing SCM into the baseline achieves a performance gain. Similarly, integrating FRM (rows 3, 4, and 5) into the baseline leads to a consistent gain in the performance. Our final approach FANet (last row), which comprises both SCM and FRM, achieves a significant improvement in performance over the baseline. The best results are in bold.}
\scalebox{1.08}{
\begin{tabular}{|l|c|c|c|c|c|} \hline
\multirow{2}{*}{Exp. No.} & \multirow{2}{*}{SCM} & \multicolumn{2}{c|}{FRM}          & \multirow{2}{*}{mIoU} & \multirow{2}{*}{Pix. Acc.} \\  \cline{3-4}
  & & High Frequency Context & Low Frequency Context &    &  \\ \hline
1. CE+ConvMLP (baseline)  &  &  &  & 38.15 & 89.26  \\  \hline
2. baseline &  \checkmark &  &  & 47.66 & 91.07  \\  \hline
3. baseline &   & \checkmark &  &  49.03 & 90.98 \\  \hline
4. baseline &   &  & \checkmark & 43.99 & 90.65  \\  \hline
5. baseline &    & \checkmark & \checkmark & 50.45 & 90.95  \\  \hline
6. FANet (Ours)  &   \checkmark & \checkmark & \checkmark & \textbf{54.89} & \textbf{91.41} \\   \hline 
\end{tabular}}
\label{tab:scm_fam_ablation_study}
\end{table*}

\subsection{Quantitative Comparison}
Tab.~\ref{tab:comparison_of_fanet_on_zerowaste} presents the comparison over  ZeroWaste-f \cite{bashkirova2022zerowaste} of our method with other state-of-the-art methods including CCT \cite{9157032}, ReCo \cite{ReCo}, DeepLabV3+ \cite{chen2017deeplab}, and FocalNet-B \cite{yang2022focalnet}.
The DeepLabV3+ \cite{chen2017deeplab} achieves a leading IoU score of 52.13\% on the   ZeroWaste-f dataset. In addition, the FocalNet-B \cite{yang2022focalnet}  is one of the best-performing models for semantic segmentation, and
also obtained mIoU of 53.26\%. We notice that our model surpasses these compared methods and obtains mIoU and pixel accuracy of 54.89\% and 91.41\%, respectively. 
Such improvements are indicative of our novel adaptive feature enhancement module's efficiency.
Our method sets a new state-of-the-art performance with this significant gain obtained in the challenging mIoU metric.

\begin{table}[t!]
\centering
\caption{Comparison of FANet with FocalNet on test set.}
\begin{tabular}{|l|c|c|c|}
\hline
\textbf{Methods} & \textbf{Params} & \textbf{mIoU} &  \textbf{Pix. Acc.} \\
\hline
FocalNet-T \cite{yang2022focalnet} & 28.6 M & 51.71 & 91.03 \\
FocalNet-B \cite{yang2022focalnet} & 88.7 M & 54.26 & {91.28} \\
\hline
FANet (ours) & 36.7 M & \textbf{54.89} &  \textbf{91.41} \\
\hline
\end{tabular}
\label{tab:effectiveness_of_FAM}
\end{table}

\begin{table}[t!]
\centering
\caption{Comparison on the validation set of ImageNet-1K.}
\begin{tabular}{|l|c|c|c|}
\hline
\textbf{Methods} & Params (M) & Flops (G) & Top1 Acc. \\
\hline
FocalNet-T & \textbf{28.6} & \textbf{4.5} & 82.3 \\
FANet (ours) & 36.7 & 5.8 & \textbf{82.5} \\
\hline
\end{tabular}
\label{tab:imagenet1k_results}
\end{table}

\subsection{Qualitative Comparison}
Fig.~\ref{fig_qualitative} illustrates the qualitative comparison of our method with DeepLabV3+ \cite{chen2017deeplab} and FocalNet-B \cite{yang2022focalnet}. We observe that our FANet can accurately segment the object in complex scenarios. For example, in the first row, our method can segment the translucent object better compared to the others. Similarly, in the second row, our method can segment objects in the presence of a heavily cluttered background. 
These results show our method's capability to adapt a variety of image characteristics and correctly classify each item which validates the robustness of our segmentation method.

\subsection{Ablation Study}
For all the ablation study experiments, the FANet backbone is pre-trained on ImageNet for 300 epochs. 

\noindent\textbf{Effectiveness of SCM and FRM:} 
Tab.~\ref{tab:scm_fam_ablation_study} showcases the impact of our contributions. Our baseline consists of a depthwise convolution embedding with a skip connection, dubbed as CE, and ConvMLP. Introducing the SCM (row-2) into the baseline which is responsible for capturing the semantic information over the larger context leads to improved performance gain compared to the baseline. On the other hand, integrating the FRM which refines the features highlighting both low-frequency and high-frequency regions also obtains a significant gain compared to the baseline as shown in rows 3, 4, and 5, respectively. 
Finally, the integration of SCM and FRM into the baseline achieves the best performance which indicates that efficiently captures the larger context information as well as preserves the low-frequency and high-frequency components. This integration improves the model's capacity to better capture the spatial descriptors as well as coarse-to-fine details, which
show the metrics of our contributions.

\noindent\textbf{Comparison of ours FANet with the FocalNet:} 
We compared our method with FocalNet-T and FocalNet-B frameworks with our method in Tab. \ref{tab:effectiveness_of_FAM}. From the table, it is clearly shown that our method exhibits better capability to handle the cluttered backgrounds. Although FocalNet-T has less the number of parameters, it exhibits a significantly reduced mIoU of 3.18\% compared to ours.
In addition, our method has significantly reduced the number of parameters compared to the FocalNet-B frameworks and still achieves better performance. This reduction in parameters signifies a more efficient model, both in terms of computational resources and processing speed, without compromising the quality of segmentation. It underscores the potential of our approach in setting new standards for segmentation tasks, particularly in complex and resource-constrained environments.

\noindent\textbf{Feature Visualization of FRM:}  
In Fig. \ref{teaser_figure_showing_FAM}, we show the effectiveness of our FRM using feature visualization. To do so, we take the stage 3 features and show the input features to our FRM, output features for both $R$ and $S$, and output of the FRM as (b), (c), (d), and (e), respectively. Whereas, (a), (f), and (g) show the input image, our prediction overlay with the input image, and ground truth overlay with the input image, respectively. From these feature visualizations, we can observe that our FRM can preserve both low-frequency and high-frequency components in the presence of a heavily cluttered environment for better semantic segmentation. This ability to preserve detailed textural information alongside broader contextual elements significantly contributes to the accuracy. 

\noindent\textbf{Comparison of FANet over ImageNet-1K:} 
Finally, in Tab.~\ref{tab:imagenet1k_results},  we compare our FANet method with FocalNet-T over the validation set of ImageNet-1K. Although our approach has more parameters, it has improved accuracy in terms of top-1. Moreover, from Tab.~\ref{tab:effectiveness_of_FAM}, it is notable that our FANet has significantly improved performance for the target ZeroWaste-f dataset compared to FocalNet-T. This shows that FocalNet-T has limited ability to tackle the cluttered background challenge for the semantic segmentation task. 

\section{Conclusion}
In this work, we propose a feature amplification network (FANet) to capture the semantic context information and generate multi-stage features for better semantic segmentation in complex scenarios especially in heavily cluttered objects. Particularly, we propose an adaptive feature enhancement (AFE) block that exploits both larger contexts using the spatial context module (SCM) and semantic cues in the feature refinement module (FRM) in a parallel fashion. Our FRM, inspired by image sharpening and contrast enhancement, exploits low-frequency and high-frequency regions in the latent space to perform feature refinement. Our extensive experimental study reveals the effectiveness of our method.



\printbibliography 
\end{document}